\documentclass[review]{elsarticle}

\usepackage{balance}  
\usepackage{graphicx} 
\usepackage{times}    
\usepackage{url}      
\usepackage{floatrow}
\usepackage{array}    
\usepackage{multirow}
\usepackage{url}

\usepackage{enumitem}
\usepackage{amsmath}
\usepackage{subcaption}
\usepackage{colortbl}
\usepackage{framed}
\usepackage{tabularx}
\usepackage{tabulary}
\usepackage{changepage}
\usepackage{pifont}
\usepackage{hyperref}

\journal{Applied Soft Computing}
\definecolor{Gray}{gray}{0.8}

\begin{document}
\begin{frontmatter}

\title{Detecting Falls with X-Factor Hidden Markov Models}


\author[one]{Shehroz S. Khan\corref{cor1}}
\ead{s255khan@uwaterloo.ca}

\author[two]{Michelle E. Karg,Dana~Kuli\'{c}}
\ead{mkarg@uwaterloo.ca}

\author[two]{Dana~Kuli\'{c}}
\ead{dkulic@uwaterloo.ca}

\author[one]{Jesse Hoey}
\ead{jhoey@uwaterloo.ca}

\cortext[cor1]{Corresponding author}

\address[one]{David R. Cheriton School of Computer Science\\
 University of Waterloo, Canada}
 
 \address[two]{Department of Electrical and Computer Engineering\\
 University of Waterloo, Canada}

\begin{abstract}
Identification of falls while performing normal activities of daily living (ADL) is important to ensure personal safety and well-being. However, falling is a short term activity that occurs rarely and infrequently. This poses a challenge for traditional supervised classification algorithms, because there may be very little training data for falls (or none at all) to build generalizable models for falls. This paper proposes an approach for the identification of falls using a wearable device in the absence of training data for falls but with plentiful data for normal ADL. We propose three `X-Factor' Hidden Markov Model (XHMMs) approaches. The XHMMs have `inflated' output covariances (observation models). To estimate the inflated covariances, we propose a novel cross validation method to remove `outliers' from the normal ADL that serves as proxies for the unseen falls and allow learning the XHMMs using only normal activities. We tested the proposed XHMM approaches on two activity recognition datasets and show high detection rates for falls in the absence of fall-specific training data. We show that the traditional method of choosing threshold based on maximum of negative of log-likelihood to identify unseen falls is ill-posed for this problem.  We also show that supervised classification methods perform poorly when very limited fall data is available during the training phase.
\end{abstract}


\end{frontmatter}


\section{Introduction}
\label{introduction}
Identification of normal Activities of Daily Living (ADL), for e.g., walking, hand washing, making breakfast, etc., is important to understand a person's behaviour, goals and actions \cite{Acampora2013}. However, in certain situations, a more challenging, useful and interesting research problem is to identify cases when an abnormal activity occurs, as it can have direct implications on the health and safety of an individual. An important abnormal activity is the occurrence of a fall. However, falls occur rarely, infrequently and unexpectedly w.r.t. the other normal ADLs and this leads to either little or no training data for them \cite{igual2013challenges}. The Centers for Disease Control and Prevention, USA \cite{website:cdc}, suggests that on average, patients incur $2.6$ falls per person per year. 
Recent studies also suggest that even in a long term experimental set up only a few real falls may be captured \cite{Debard2012Camera,Stone2015Fall}. 
In these situations with highly skewed fall data, a typical supervised activity recognition system may misclassify `fall' as one of the already existing normal activity as `fall' may not be included in the classifier training set. An alternative strategy is to build fall detection specific classifiers that assume abundant training data for falls, which is hard to obtain in practice. Another challenge is the data collection for falls, as it may require a person to actually undergo falling which may be harmful, ethically questionable, and the falling incidences collected in controlled laboratory settings may not be the true representative of falls in naturalistic settings \cite{kangas2012comparison}. 
 
The research question we address in this paper is: \textit{Can we recognise falls by observing only normal ADL with no training data for falls in a person independent manner?}. We use the HMMs for the present task as they are very well-suited for sequential data and can model human motions with high accuracy \cite{kulic08}. 
Typically, an HMM can be trained on normal activities and the maximum of negative of log-likelihood on the training data is set as a threshold to identify a fall as an outlier. However, choosing such a threshold may severely effect classifier's performance due to spurious artifacts present in the sensor data and most of the falls may be classified as normal activities. In this paper, we use the outlier detection approach to identify falls and present three X-Factor HMM based approaches for detecting short-term fall events. The first and second method models individual normal activities by separate HMMs or all normal activities together by a single HMM, by explicitly modelling the poses of a movement by each HMM state. An alternative HMM is constructed whose model parameters are the averages of the normal activity models, while the averaged covariance matrix is artificially `inflated' to model unseen falls. In the third method, an HMM is trained to model the transitions between normal activities, where each hidden state represents a normal activity, and adds a single hidden state (for unseen falls) with an inflated covariance based on the average of covariances of all the other states. The inflation parameters of the proposed approaches are estimated using a novel cross-validation approach in which the outliers in the normal data are used as proxies for unseen fall data. 
We present another method that leverages these outliers to train a separate HMM as a proxy model to detect falls. 
We also compare the performance of one-class SVM and one-class nearest neighbour approach along with several supervised classification algorithms that use full data for normal activities but the number of falls are gradually increased in the training set. We show that supervised classifiers perform worse when limited data for falls is available during training.
This paper is a comprehensive extension of the work of Khan et al. \cite{khan2014iwaal} in terms of :
\begin{itemize}
  \item Proposing two new models to detect unseen falls by (i) modelling transitions among normal activities to train an HMM and adding a new state to model unseen falls, and (ii) training a separate HMM on only the outliers in the normal activities data to model unseen falls.
  \item Data pre-processing, extraction of signals from raw sensor data, and number and type of features are different from Khan et al.\cite{khan2014iwaal}.
  \item Studying the effect of changing the number of states on the proposed HMM methods for fall detection.
  \item Identifying similarity through experiments between the rejected outliers from the normal activities and the unseen falls.
  \item Additional experiments evaluating the effect of quantity of fall data available during the training phase on the performance of the supervised versions of the proposed fall detection methods and two other supervised classification methods.
 \end{itemize}


\section{Related Work}
\label{related_work}
The research in fall detection spans over two decades with several recent papers \cite{igual2013challenges,Mubashir:2013:SFD,kwolek2014human} that discuss different methodologies, trends and ensuing challenges using body worn, ambient or vision based fall detection techniques. 
Several research works in fall detection are based on thresholding techniques \cite{Bourke200884} or supervised classification \cite{igual2013challenges}. One of the major challenges in fall detection is the less availability of fall data \cite{Stone2015Fall}; therefore, such techniques are difficult to use in practice. Keeping this view in mind, we survey techniques that attempt to detect falls by employing generative models, outlier/anomaly detection and one-class classification \cite{Khan:KER:2014} based techniques that only use data from normal activities to build the model and identify a fall as an anomaly or outlier.

Thome et al. \cite{DBLP:conf/icarcv/ThomeM06} present a Hierarchical HMM (HHMM) approach for fall detection in video sequences. The HHMMs first layer has two states, an upright standing pose and lying. They study the relationship between angles in the 3D world and their projection onto the image plane and derive an error angle introduced by the image formation process for a standing posture. Based on this information, they differentiate other poses as `non-standing' and thus falls can be distinguished from other motions. A two-layer HMM approach, \textit{SensFall} \cite{Luo:Liu:Liu:2012}, is used to identify falls from other normal activities. In the first layer, the HMM classifies an unknown activity as normal vertical activity or `other', while in second stage the `other' activity is classified as either normal horizontal activity or as a fall. 
Tokumitsu et al. \cite{Tokumitsu:2011} present an adaptive sensor network intrusion detection approach by human activity profiling. They use multiple HMMs for every subject in order to improve the detection accuracy and consider the fact that a single person can have multiple patterns for the same activity. The data is collected using infra-red sensors. A new sequence of activity is fed to all the HMMs and likelihoods are computed. If all the likelihoods calculated from corresponding HMMs are not greater than pre-determined thresholds, then an anomaly is identified. 
Cheng et al. \cite{iet/conferences/chen:luo:2011} present a fall detection algorithm based on pattern recognition and human posture analysis. The data is collected through tri-axial accelerometer embedded in the smartphones and several temporal features are computed. HMM is employed to filter out noisy character data and to perform dimensionality reduction. One-class SVM (OSVM) is applied to reduce false positives, followed by a posture analysis to counteract the missed alarms until a desired accuracy is achieved. 

Zhang et al. \cite{hunag:li:irwin:2006} trained  an OSVM from positive samples (falls) and outliers from non-fall ADL and show that the falls can be detected effectively. Yu et al. \cite{Miao:Naqvi:2011} propose to train Fuzzy OSVM on fall activity captured using video cameras and to tune parameters using fall and some non-fall activities. Their method assigns fuzzy membership to different training samples to reflect their importance during classification and is shown to perform better than OSVM. 
Popescu \cite{Popescu2009} presents a fall detection technique that uses acoustic signals of normal activities for training and detects fall sounds from it. They train OSVM, one-class nearest neighbour (OCNN) classifier and One-class GMM classifier (that uses a threshold)  to train models on normal acoustic signals and find that OSVM performs the best; however, it is outperformed by its supervised counterpart. Medrano et al. \cite{medrano2014detecting} propose to identify falls using a smartphone as a novelty from the normal activities and found that OCNN performs better than OSVM but is outperformed by supervised SVM. 

The supervised and thresholding techniques for fall detection collect artificial fall data in a laboratory under non-naturalistic settings; however, such fall data may not be true representative of actual falls and learning with them may lead to over-fitting.
To overcome the need for a sufficient set of representative `fall' samples, we propose three `X-Factor' HMM based approaches to identify falls across different people while learning models only on data from normal activities.

\section{Proposed Fall Detection Approaches}
\label{sec:proposed}

The problem we investigate in this paper pertains to activity recognition and the datasets we use capture the temporal activities performed by humans. 
The Hidden Markov Models (HMM) are effective in  modelling the temporal dynamics in data sequences and consider the history of actions when taking a decision on the current sequence. The HMM is a doubly stochastic process for modelling generative sequences that can be characterized by an underlying process generating an observable sequence. Formally, an HMM  consists of the following components \cite{rabiner}:
\begin{itemize}
  \item \textbf{N} -- the number of hidden states in the HMM. The hidden states can be connected in several ways, for example in left-to-right manner or fully interconnected (ergodic). the set of states can be denoted as $S=\{S_1,S_2,\ldots,S_N\}$ and the state at time $t$ as $q_t$.

  \item \textbf{M} -- The number of distinct observation symbols per state that corresponds to the physical output of the system being modelled. The symbols can be denoted as $V=\{v_1,v_2,\ldots,v_M\}$. When the observation is continuous, $M=\infty$, and can be approximated using Gaussian or mixture of Gaussian with mean and covariance corresponding to each hidden state as the underlying parameters.

  \item \textbf{A} -- The state transition probability distribution $A=a_{ij}$, where $a_{ij}$ represents the probability of state $j$ following state $i$ and is expressed as:
  \begin{equation}
      a_{ij}=P[q_{t+1}=S_j|q_t=S_i]\quad\quad 1 \le i,j \le N   
  \end{equation}
  
  The coefficients of state transition have the following properties:
  
  \[a_{ij} \ge 0,\quad \sum_{j=1}^N a_{ij}=1\]
  
  The state transition matrix $A$ is independent of time. For the ergodic design where any state can reach any other state $a_{ij}>0$ for all $i$ and $j$, whereas for other topologies one or more values will have $a_{ij}=0$.
  \item \textbf{B} -- The observation symbol probability distribution in state $j$, $B=\{b_j(k)\}$, where
  \begin{equation}
    b_j(k)=P[v_k\quad at\quad t|q_t=S_j] \quad\quad 1 \le j \le N, 1 \le k \le M
  \end{equation} 
  
  \item \textbf{$\pi$} -- The initial state distribution $\pi = \{\pi_i\}$, where 
  \begin{equation}
    \pi_i=P[q_1=S_i]\quad\quad 1 \le i \le N
  \end{equation}
\end{itemize}
  
The complete set of parameters of an HMM can also be compactly represented as \cite{rabiner}:
  \begin{equation}
    \lambda =(\pi,A,B)
  \end{equation}

\begin{figure}[H]
  \includegraphics[scale=0.4]{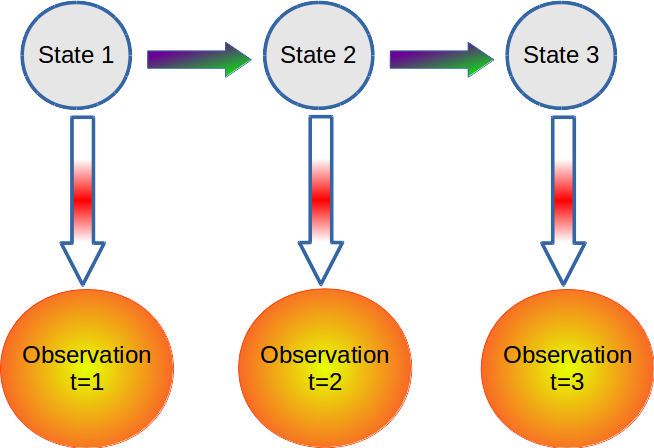}
  \caption{Discrete HMM with $3$ states and $3$ possible outputs}
  \label{fig:hmm}
\end{figure}

A pictorial representation of a $3$ state discrete HMM is shown in Figure \ref{fig:hmm}. The model follows a Markovian assumption, i.e., the current state at time $t$ is independent of all states $t-2,\ldots, 1$ given the state at $t-1$ and an independence assumption, i.e., the output observation at time $t$ is independent of all the previous observations and states given the current state. 

HMMs are successfully used in detection of human activities with high accuracy \cite{kulic08}. Mannini and Sabatini \cite{Manninis100201154} compare various single-frame classifiers against HMM based sequential classifier for activity recognition using on-body accelerometers and report superior performance of the HMM classifiers.
Typically, two approaches are commonly applied to model human actions and activities using HMMs \cite{kulic08}:  
\begin{enumerate}[label=(\roman*)]
  \item \textit{Modelling Poses}: Train an HMM for an activity by explicitly modelling the poses of a movement by each state, or  
  \item \textit{Modelling Activities}: Train an HMM for different activities by modelling each activity by a single state.
\end{enumerate}

We consider both of these approaches to propose `X-Factor' based models to identify falls when their training data is not available, which is discussed next.

\subsection{Pose HMM}
\label{sec:pose}
The traditional method to detect unseen abnormal activities is to model each normal activity using an HMM (by modelling the poses of a movement by each state), compare the likelihood of a test sequence with each of the trained models and if it is below a pre-defined threshold for all the models then identify it as an anomalous activity \cite{khan2014iwaal}. For fall detection, we model each normal activity $i$ by an ergodic HMM which evolves through a number of $k$ states. The observations $o_j(t)$ in state $j$ are modelled by a single Gaussian distribution. 
Each model $i$ is described by the set of parameters, $\lambda_i = \{\pi_i, A_i,(\mu_{ij}, \Sigma_{ij}) \}$, where $\pi_i$ is the prior, $A_i$ is the transition matrix, and $\mu_{ij}$ and $\Sigma_{ij}$ are the mean and covariance matrix of a single Gaussian distribution, $\mathcal{N}(\mu_{ij},\Sigma_{ij})$, giving the observation probability $Pr(o_i|j)$ for the $j^{th}$ HMM state.  
This method estimates the probability that an observed sequence has been generated by each of the $i$ models of normal activities. If this probability falls below a threshold $T_i$ for each HMM, a fall is detected. Typically, an HMM is trained for each normal activity on the full training data and the individual activity threshold is set as the maximum of the negative log-likelihood of the training sequences  (we call this method as $HMM1$). If a new activity's negative log-likelihood is below each of these thresholds, it is identified as a fall. 

Quinn et al. \cite{quinn} present a general framework based on Switched Linear Dynamical Systems for condition monitoring of a premature baby receiving intensive care by introducing the `X-factor' to deal with unmodelled variation from the normal events that may not have been seen previously. This is achieved by inflating the system noise covariance of the normal dynamics to determine the regions with highest likelihood which are far away from normality based on which events can be classified as `not normal'. We extend this idea to formulate an alternate HMM (we call this approach as $XHMM1$) to model unseen fall events. This approach constructs an alternate HMM to model fall events by averaging the parameters of $i$ HMMs and increasing the averaged covariances by a factor of $\xi$ such that each state's covariance matrix is expanded. Thus, the parameters of the X-Factor HMM will be $\lambda_{XHMM1} = \{\bar{\pi}, \bar{A}, \bar{\mu}, \xi \bar{\Sigma}) \}$, where $\bar{\pi}$, $\bar{A}$, $\bar{\mu}$, and $\bar{\Sigma}$ are the average of the parameters $\pi_i$, $A_i$, $\mu_i$ and $\Sigma_i$ of each $i$ HMMs. Each of the $i$ HMMs is trained on non-fall data obtained after removing outliers from the normal activities and these outliers serve as the validation set for optimizing the value of $\xi$ using cross validation (see details in Section \ref{sec:threshold}). For a test sequence, the log-likelihood is computed for all the HMM models ($i$ HMMs representing $i$ normal activities and the alternate HMM representing fall events) and the one with the largest value is designated as its class label.

\subsection{Normal Pose HMM}
\label{sec:normal_pose}
Another method to identify abnormal activities is to model all the normal activities together using a single HMM and if a test sequence's likelihood falls below a predefined threshold, it is identified as anomalous \cite{Khan:2012:TDU:2370216.2370444}. For fall detection, we group all the normal activities together and train a single HMM; where normal poses are modelled by each state. The idea is to learn the `normal concept' from the labelled data. This method estimates the probability that the observed sequence has been generated by this common model for all the normal activities and if this probability falls below a threshold $T$, a fall is detected. Typically the maximum of negative log-likelihood on the training data is set as a threshold to detect unseen falls (we call this method $HMM2$). 
Similar to $XHMM1$, we propose to construct an alternative HMM to model the `fall' activities whose parameters ($\lambda_{XHMM2}$) remain the same as the HMM to model non-fall activities together ($\lambda$) except for the covariance, whose inflated value is computed using cross validation (we call this method ($XHMM2$); see details in Section \ref{sec:threshold}). For a test sequence, the log-likelihood is computed for both HMM models (HMM representing non-fall activities and the alternate HMM representing fall events) and the one with the larger value is designated as its class label.

The intuition behind $XHMM1$ and $XHMM2$ approaches is that if the states representing non-fall activities are modelled using Gaussian distributions, then the fall events coming from another distribution can be modelled using a new Gaussian (X-factor) with larger spread but with the same mean as non-fall activities. 
The observations that are closer to the mean retain high likelihood under the original Gaussian distribution for the normal activities, whereas the X-factor will have higher likelihood for observations that are far away from the normal activities. To simplify the assumptions about unseen falls, other extra factors such as the mean and the number of states are not introduced in the proposed approaches.

\subsection{Activity HMM}
\label{sec:xhmm3}
Smyth \cite{smyth1994markov} addresses the problem of real-time fault monitoring, where it is difficult to model all the unseen fault states of a system and proposes to add a $(j+1)$ novel hidden state (in an HMM) to cover all other possible states not accounted by the known $j$ states. The novel state's prior probability is kept same as other known states and the density of the observable data given the unknown state is defined by using non-informative Bayesian priors. 
For detecting falls, we train a single HMM to model transitions of normal activity sequences, with parameters, $\lambda_{XHMM3}=\{\pi,A,\mu,\Sigma\}$, where each hidden state represents a normal activity, and add an extra hidden state to the model; its means and covariances are estimated by averaging the means and covariances of all other states representing the normal activities. The X-factor is introduced to vary the covariance of this novel state by a factor of $\xi$, which can be determined using cross validation (see Section \ref{sec:threshold}). Adding a novel state to the existing HMM means adding a row and column to $A$ to represent transitions to and from the state capturing unseen fall. However, this information is not available apriori. For fault detection application, Smyth \cite{smyth1994markov} designs a $3$ state HMM and added a novel $4^{th}$ state to model unknown anomalies and chooses the probability of remaining in the same state as $0.97$ and distributes transition to other states uniformly.  We use similar idea to choose probability of $0.95$ to self transitions to fall events and the rest of the probability is uniformly distributed for transitions from fall events to normal activities. For transitions from different normal activities to falls, a probability of $0.05$ is set (to capture the assumption that falls occur rarely) and the transition probabilities between different normal activities are scaled such that the total probability per row in the matriix $A$ sums up to $1$. 
Viterbi decoding \cite{rabiner} is employed on a test sequence to find the most likely hidden state that generated it, if it consists of the novel state, the sequence is classified as a fall or else a normal activity.


\subsection{$HMM_{NormOut}$}
As discussed in Section \ref{sec:pose} and \ref{sec:normal_pose}, some outliers are rejected from each of the normal activities that may arise due to artifacts in the sensor readings or mislabelling of training data. These rejected sensor readings from each normal activity are grouped together and two HMMs are trained, one each for non-fall activities and outlier activities. We call this approach as $HMM_{NormOut}$. The  HMM model learnt on outliers activities may not be the true representative for falls but it can model those activities that are non-falls.

 \section{Threshold Selection and Proxy Outliers}
\label{sec:threshold}
As discussed in Section \ref{introduction}, falls occur rarely and infrequently compared to normal activities; therefore, it is difficult to get labelled data for them. This may result in situations with abundant data for normal activities and none for falls. 
To detect falls using traditional HMM approaches ($HMM1$ and $HMM2$), typically, a threshold is set on the likelihood of the data given an HMM trained on this `normal' data. This threshold is normally chosen as the maximum of negative log-likelihood \cite{Khan:2012:TDU:2370216.2370444}, and can be interpreted as a slider between raising false alarms or risking missed alarms \cite{Tokumitsu:2011}. A major drawback of this approach is that it assumes that the data for each normal activity is correctly labelled and sensor readings are non-spurious. This assumption can be detrimental for fall detection performance; any abnormal sensor reading or mislabelling of training data can alter this threshold and adversely effect the performance. For the proposed approaches, another challenge is to estimate the parameter $\xi$ for $XHMM1$,  $XHMM2$ and  $XHMM3$ in the absence of fall data during the training phase. 

To address the above mentioned issues and finding appropriate $\xi$, we propose to use the deviant sequences (\textit{outliers}) within the `normal' data. The idea is that even though the `normal' data may not contain any falls, it may contain sensor readings that are spurious, incorrectly labelled or significantly different.  These outliers can be used to set $\xi$ that are required for fall detection, thereby serving as a proxy for the fall data in order to learn the parameter $\xi$ of the three XHMMs.  To find the outliers,  we use the concept of quartiles of a ranked set of data values that are the three points that divide the data set into four equal groups, where each group comprises of a quarter of the data. Given the log-likelihoods of sequences of training data for an HMM and the lower quartile ($Q_1$), the upper quartile ($Q_3$) and the inter-quartile range ($IQR = Q_3 - Q_1$), a point $P$ is qualified as an outlier if
\begin{equation}
  P > Q_3 + \omega \times IQR \quad || \quad P < Q_1 - \omega \times IQR 
\label{eqn:iqr}
\end{equation}
where $\omega$ represents the percentage of data points that are within the non-extreme limits. Based on $\omega$, the extreme values of log-likelihood that represent spurious training data can be removed, that leads to the 
\begin{enumerate}[leftmargin=*]
  \item Creation of a validation set comprising of outliers (proxies for falls), and 
  \item Computation of parameter $\xi$ for the proposed $XHMM$ approaches.
\end{enumerate}

\begin{figure}[!ht]
  \centering
  \includegraphics[width=\linewidth, height=3.5cm]{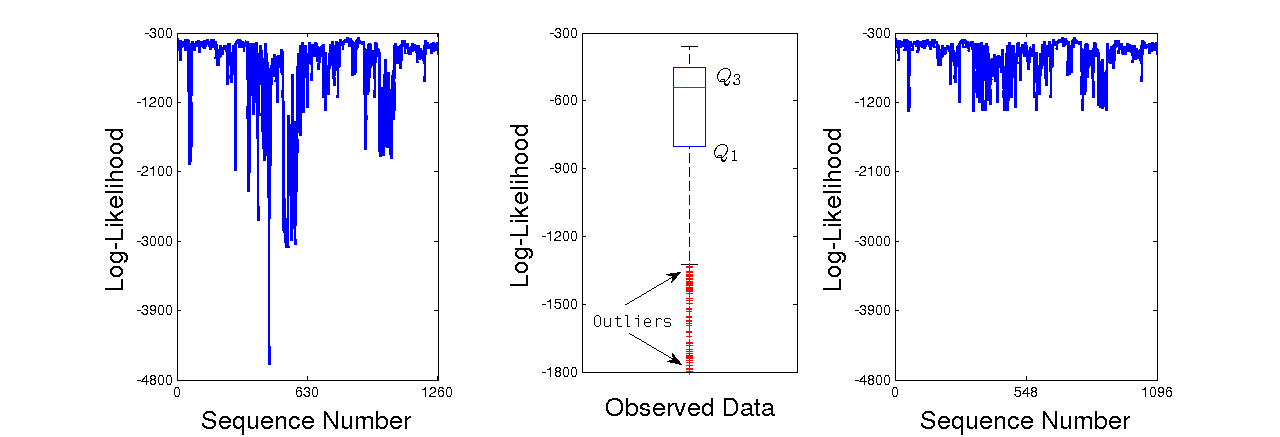}
  (a)~~~~~~~~~~~~~~~~~~~~~~~~~(b)~~~~~~~~~~~~~~~~~~~~~~~~~(c)    
  \caption{Log-Likelihoods (a) before and (c) after outlier removal.  (b) shows box-plot of the quartiles for this data and the outliers for $w=1.5$}
  \label{fig:box-plot}
\end{figure}

Figure \ref{fig:box-plot} (a) shows the log-likelihood $\log Pr(O|\lambda_{running})$ for $1262$ equal length ($1.28$ seconds) running activity sequences of the DLR dataset (see Section \ref{sec:dataset}). Figure~\ref{fig:box-plot} (b) shows a box plot with the quartiles and the outliers (shown as {\color{red} +}) for $w=1.5$.  Figure~\ref{fig:box-plot} (c) shows the same data as in Figure~\ref{fig:box-plot}(a) but with the outliers removed. 

\begin{figure}[ht]
\vspace{-4mm}
\centering
  \includegraphics[trim=0cm 14cm 1cm 8cm, clip, width=7cm, height=4cm]{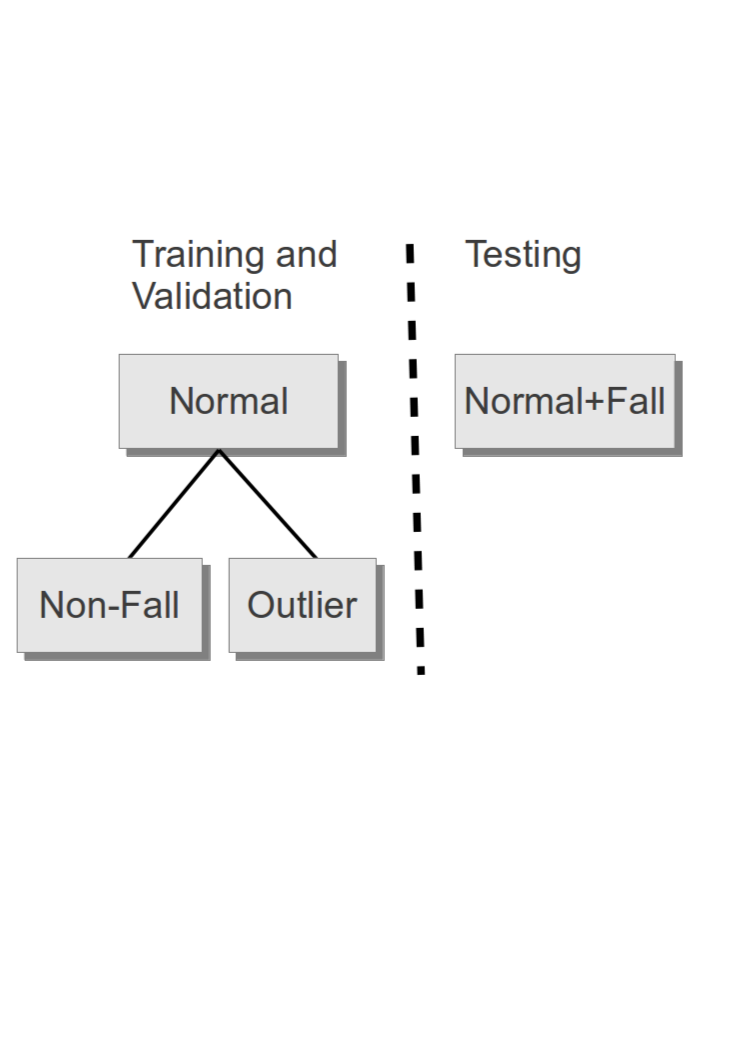}
\caption{Cross Validation Scheme}
\label{fig:cv-diag}
\vspace{-2mm}
\end{figure}

We employ an internal cross-validation to train the three XHMMs using only the non-fall data. We first split the normal data into two sets: `non-fall' data and `outlier' data (see Figure~\ref{fig:cv-diag}).  We do this using Equation~\ref{eqn:iqr} with a parameter $\omega$ that is manually set and only used for this initial split. For each activity, an HMM is trained on full normal data and based on $\omega$, `outliers' are rejected from them and the remaining data is considered as `non-fall'. To optimize the covariance parameter, $\xi$, we use a $K$-fold cross validation: the HMMs are trained on $\left(\frac{K-1}{K}\right)^{th}$ of the  `non-fall' data, and tested on $\left(\frac{1}{K}\right)^{th}$ of the  `non-fall' data and on all the `outlier' data.  This is done $K$ times and repeated for different values of $\xi$. The value of $\xi$ that gives the best averaged performance metric (see Section \ref{sec:perfmet}) over $K$-folds is chosen as the best parameter. Then, each classifier is re-trained with this value of parameter on the `non-fall' activities. 

\section{Experimental Design}
\label{experiments}
\subsection{Datasets}
\label{sec:dataset}
The proposed fall detection approaches are evaluated on the following two human activity recognition datasets.
\begin{enumerate}[leftmargin=*] 
  \item German Aerospace Center (DLR) \cite{dlr65511}: This dataset is collected using an Inertial Measurement Unit with integrated accelerometer, gyroscope and 3D magnetometers with sampling frequency of $100$ Hz. The dataset contains samples taken from $19$ people under semi-natural conditions. The sensor was placed on the belt either on the right/left side of the body or in the right pocket in different orientations. The dataset contains $7$ activities: standing, sitting, lying, walking (up/downstairs, horizontal), running/jogging, jumping and falling. One subject did not perform fall activity and its data is omitted from the analysis. 
  
  \item MobiFall (MF) \cite{mobifall}: This dataset is collected using a Samsung Galaxy S3 device equipped with $3$D accelerometer and gyroscope. The mobile device was placed in a trouser pocket in random orientations. Mean sampling of $87$ Hz is reported for accelerometer and $200$ Hz for the gyroscope. The dataset is collected from $11$ subjects; eight normal activities are recorded in this dataset: step-in car, step-out car, jogging, jumping, sitting, standing, stairs (up and down joined together) and walking. Four different types of falls are recorded -- forward lying, front knees lying, sideward lying and back sitting chair. Different types of falls are joined together to make one separate class for falls. Two subjects only performed fall activity and their data is removed from the analysis.
\end{enumerate}

 The DLR dataset is collected in semi-naturalistic settings; therefore, the ratio of falls to normal activities is quite small $\approx 0.0032$ ($26576$ normal activities segments and $84$ fall segments), whereas in the MF dataset this ratio is $\approx 0.0899$ ($5430$ normal activities and $488$ fall segments). 

\subsection{Data Pre-Processing}
For the MF dataset, the gyroscope sensor has a different sampling frequency than the accelerometer and their time-stamps are also not synchronized; therefore, the gyroscope readings are interpolated to synchronize them with the accelerometer readings. Although the calibration matrix for the DLR data is available to rotate the sensor readings to the world frame, in our experiments we did not use it because it did not improve the results. For the MF dataset, orientation information is present but incorporating it led to the deterioration of results. This observation is consistent with the work of de la Vega et al. \cite{delavega2013} that suggest that activities can be detected without considering the orientations. Winter \cite{winter2009biomechanics} suggests that for the walking activity, $99.7\%$ of the signal power was contained in the lower seven harmonics (below $6$Hz), with evidence of higher-frequency components extending up to the $20^{th}$ harmonic. Beyond that frequency, the signal had the characteristics of `noise', which can arise from different sources, such as electronic/sensor noise, spatial precision of the digitization process, and human errors. Therefore, for both the datasets, the sensor noise is removed by using a $1^{st}$ order Butterworth low-pass filter with a cutoff frequency of $20$Hz. 
  The signals are segmented with $50\%$ overlapping windows, where each window size is $1.28$ seconds for DLR dataset and $3$ seconds for MF dataset to simulate a real-time scenario with fast response. The reason that DLR dataset does not have the same windows size as MF dataset is that it contains short duration fall events. Therefore, when the window size is increased to $3$ seconds, fall samples could not be extracted for many subjects and cross-validation across different subjects (see Section \ref{sec:perfmet}) may not work as desired.

\subsection{Feature Extraction}
\label{sec:feature_extraction}
The literature on feature extraction from motion sensors is very rich \cite{Ravi:2005,huynh2005analyzing,cruz2013features}. Most of the feature extraction techniques involve computing time domain, frequency domain, and statistical features from the sensor readings.
We extract the following five signals from each of the datasets:

\begin{enumerate}[leftmargin=*]
  \item Three acceleration readings $a_x, a_y,a_z$ along the $x$, $y$ and $z$ directions,
  \item Norm of acceleration, $a_{norm} = \sqrt{a_x^2+a_y^2+a_z^2}$ and gyroscope, $\omega_{norm} = \sqrt{\omega_x^2+\omega_y^2+\omega_z^2}$, where $\omega_x$, $\omega_y$ and $\omega_z$ are the angular velocities in the $x$, $y$ or $z$ direction.
  
\end{enumerate}

Considering three separate acceleration signals will be useful in obtaining direction specific information, whereas the norm of acceleration and gyroscope will be useful in extracting orientation-invariant information.
One objective of this study is to identify low-cost features that are highly discriminative in identifying various types of normal activities. Therefore, we extract $31$ standard time and frequency domain features from these signals (as shown in Table \ref{tab:features} along with their description). 
Features are computed for each window for $XHMM3$. To extract temporal dynamics for $HMM1$, $HMM2$, $XHMM1$, $XHMM2$ and $HMM_{NormOut}$, each window is sub-divided into $16$ms frames and features are computed for each frame. 

\begin{table}
\vspace{-15mm}
  \centering
 \begin{adjustwidth}{-2cm}{}
\begin{tabular}{| >{\centering\arraybackslash}m{1.3cm}|p{6.9cm}|m{6cm}| } \hline
    \textbf{\#Features}  & \textbf{Type of feature}& \textbf{Reason to Use} \\ \hline
    $f_1$ -- $f_5$       & Mean of $a_x, a_y,a_z, a_{norm}, \omega_{norm}$ \cite{dlr65511} & Average features are used for the detection of body positions \cite{Ermes2008}. These feature work well in identifying various ADL \cite{dlr65511}\\ \hline
    $f_6$ -- $f_{10}$    & Maximum value of  $a_x, a_y,a_z, a_{norm}, \omega_{norm}$ \cite{dlr65511}  & These feature work well in identifying various ADL \cite{dlr65511}\\ \hline
    $f_{11}$ -- $f_{15}$ & Minimum value of  $a_x, a_y,a_z, a_{norm}, \omega_{norm}$ \cite{dlr65511} & These feature work well in identifying various ADL \cite{dlr65511}\\ \hline
    $f_{16}$ -- $f_{20}$ & Standard Deviation of $a_x, a_y,a_z, a_{norm}, \omega_{norm}$ \cite{dlr65511} & Variance feature is used for estimating the intensity of an activity \cite{Ermes2008}. These feature work well in identifying various ADL \cite{dlr65511}\\ \hline
    $f_{21}$ -- $f_{22}$ & IQR of $a_{norm}, \omega_{norm}$ \cite{dlr65511}  & These feature work well in identifying various ADL \cite{dlr65511}\\ \hline
    $f_{23}$             & Normalized Signal Magnitude Area  \cite{khan2008accelerometer}& This is useful to identify dynamic and static activities, e.g., running or walking versus lying or standing.\\ \hline
    $f_{24}$             & Normalized Average Power Spectral Density of $a_{norm}$ & This feature is useful for the detection of cyclic activities, e.g., walking, running, cycling \cite{Ermes2008}. \\ \hline
    $f_{25}$             & Spectral Entropy of $a_{norm}$ \cite{Ermes2008} & This is useful for differentiating between activities involving locomotion.\\ \hline
    $f_{26}$             & DC component after FFT of $a_{norm}$ \cite{DBLP:conf/pervasive/BaoI04} & The is shown to result in accurate recognition of certain postures and activities. \\ \hline
    $f_{27}$             & Energy, i.e., sum of the squared discrete FFT component magnitudes of $a_{norm}$ \cite{DBLP:conf/pervasive/BaoI04}& This is shown to result in accurate recognition of certain postures and activities.\\ \hline
    $f_{28}$             & Normalized Information Entropy of the Discrete FFT component magnitudes of $a_{norm}$ \cite{DBLP:conf/pervasive/BaoI04} & This helps in discriminating activities with different energy values.\\ \hline
    $f_{29}$ -- $f_{31}$   & Correlation between $a_x, a_y, a_z$ \cite{cleland2013optimal} & This helps in differentiating among activities that involve translation in one dimension, e.g., walking and jogging from taking the stairs up and down. \\ \hline
    \end{tabular}
    \captionof{table}{Extracted Features and their Description.}
    \label{tab:features}
\end{adjustwidth}
\end{table}

\subsection{HMM Modelling}   
For all the HMM based fall detection methods discussed in the paper, the observation model uses single Gaussian distribution, diagonal covariance matrix is used for each of the HMMs and the upper and lower values are constraint to $100$ and $0.01$ during the training.  For optimizing the parameters $\xi$, a $3$-fold internal cross validation is used.
For all the HMMs methods except $XHMM3$, the following procedure is adopted:
 \begin{itemize}[leftmargin=*]
 \item Each activity in the HMMs is modelled with $2/4/8$ states, where each individual state represents functional phases of the gait cycle \cite{karg2014clinical} or the `key poses' of each activity.
 \item Five representative sequences per activity are manually chosen to initialize the parameters. Initialization is done by segmenting a single sequence into equal parts (corresponding to the number of states) and computing  $\mu_{ij}$ and $\Sigma_{ij}$ for each part and further smoothing by BW with $3$ iterations. 
 \item The transition matrix $A_i$ is ergodic (i.e. every state has transitions to other states) and initialized such that transition probabilities from one state to another are $0.025$, self-transitions are set accordingly \cite{smyth1994markov}, and the actual values are learned by BW algorithm following initialization. 
 \item The prior probabilities of each state, $\pi$, are initialized to be uniformly distributed (to sum across all states to $1$) and further learned during BW. 
 \item The likelihood for a test sequence is computed using the forward algorithm \cite{rabiner} and the classification decisions are taken based on them.

 \end{itemize}
 
 For $XHMM3$, the parameters $\mu_{j}$ and $\Sigma_{j}$ and transition matrix are computed from the annotated data and no additional BW step is used. When a  novel state is added, its parameters are estimated by averaging the means and covariances of all other states (with covariance further inflated using X-Factor) and transition matrix is re-adjusted (refer to Section \ref{sec:xhmm3}). The prior probabilities of each state is kept uniform. The decision to detect a fall is taken using the Viterbi algorithm \cite{rabiner}, which finds the most likely hidden state that produces the given observation. 
 

\subsection{Performance Evaluation and Metric}
\label{sec:perfmet}
To evaluate the performance of the proposed approaches for fall detection, we perform leave-one-subject-out cross validation (LOOCV) \cite{he2009activity}, where \emph{only} normal activities from $(N-1)$ subjects are used for training and the $N^{th}$ subject's normal activities and falls are used for testing. This process is repeated $N$ times and the average performance metric is reported. This evaluation is person independent and demonstrates the generalization capabilities as the subject who is being tested is not included in training the classifiers.  The different values of $\xi$ used in internal cross validation for $XHMM1$, $XHMM2$ and $XHMM3$ are $[1.5, 5, 10, 100]$. The value of $\omega$ is set to $1.5$ for obtaining outliers from the normal activities.

Conventional performance metrics such as accuracy, precision, recall, etc., may not be very useful when classifiers are expected to observe a skewed distribution of fall events w.r.t. normal activities.  We use the geometric mean ($gmean$) \cite{kubat1997addressing} as the performance metrics because it measures the accuracies separately on each class, i.e., it combines True Positive Rates ($TPR$) and True Negative Rates ($TNR$) and is given by $gmean=\sqrt{TPR*TNR}$. An important property of $gmean$ is that it is independent of the distribution of positive and negative samples in the test data. We also use two other performance metrics, fall detection rate ($FDR$) (or the true positives) and false alarm rate ($FAR$) (or the false positives) to better understand the performance of the proposed fall detection classifiers. A fall detection method that gives high $gmean$, high $FDR$ and low $FAR$ is considered to be better than others. 

 
%
\section{Results}
\label{results}
In this section we present the fall detection results using the DLR and MF datasets. In the first experiment, the models are learned using only the normal activities and falls are shown during testing phase only. In the second experiment, we assume the presence of few falls in the training set to build supervised models on both falls and normal activities and test the performance of these models. In the third experiment, we test our hypothesis that outliers from normal activities are similar to falls or not.

\subsection{Training without fall data}
In this experiment, we compare the performance of the fall detection methods discussed in Section \ref{sec:proposed}. $HMM1$ and $HMM2$ are trained on full `normal' data, while the proposed three $XHMM$s are trained on `non-fall' data, but they make use of full `normal' data to optimize their respective parameters. 
We also compare the results with One-Class SVM (OSVM) \cite{Khan:KER:2014} and One-class nearest neighbour (OCNN)  \cite{khan2010kernels} that are trained on only the full `normal' data. The OSVM method has an built-in mechanism to reject fraction of positive samples ($\nu$) to help deciding the class boundary in the absence of data from the negative class. We set this parameter to a default value of $\nu=0.5$ and implemented OSVM using MATLAB \cite{website:fitcsvm}. The OSVM uses a gaussian kernel by default for one-class learning. For OCNN, we keep the value of k-nearest neighbours to be $1$. 
For the HMM based methods, except for $XHMM3$ where the number of states equals the number of labelled normal activities plus an additional state for modelling falls, the number of states are varied for all other fall detection methods to study the change in performance by increasing the complexity of the models. The number of states tested are $2$, $4$ and $8$ for both the data sets. We observe that increasing the number of states do not significantly improve the performance of any methods. Though large number of states increase the training time for the models significantly. For a given fixed length sequence (for both the DLR and COV datasets), training a $8$ state HMM takes almost two times longer than a $4$ state HMM, which in turn takes almost twice to train a $2$ state HMM. We choose $4$ states HMM as the optimum for this and subsequent experiments because it provides a good trade-off between accuracy and running time. 
   
   Tables \ref{tab:exp1b} shows the performance of the different fall detection methods in the absence of training data for falls on both the datasets. We observe that for both the DLR and MF datasets, $HMM1$ and $HMM2$ failed to detect any (or most of the) falls. For DLR dataset, $XHMM3$, and $XHMM1$ show the highest $gmean$ in comparison to other methods. $HMM_{NormOut}$ performs worse than the three $XHMM$s  but better than $HMM$s. $XHMM2$ has the highest $FDR$ but at the cost of high $FAR$. Both OCNN and OSVM perform worse than the proposed XHMM methods. OCNN identified most of the falls at the cost of large number of false alarms and OSVM missed to detect most of the falls. 
   For the MF dataset, $XHMM2$ performs the best, $XHMM1$ and $XHMM3$ did not perform well because they classify most falls as step-in car and sitting. The reason for their poor performance is that the fall signals collected in this dataset contain sensor readings after the subject has hit the ground. 
Therefore, the fall data has some stationary values after the falling action has occurred. After creating overlapping windows, some of them may contain stationary values that are likely to be classified as one of the static activities. OCNN and OSVM perform worse with high falls detection rate but with large false alarms rate.

\begin{table}[!ht]
\centering
\begin{tabular}{| >{\centering\arraybackslash}m{2.1cm}| >{\centering\arraybackslash}m{0.9cm}| >{\centering\arraybackslash}m{0.65cm}| >{\centering\arraybackslash}m{0.7cm}|| >{\centering\arraybackslash}m{0.9cm}| >{\centering\arraybackslash}m{0.65cm}| >{\centering\arraybackslash}m{0.7cm}|} \hline
\multirow{2}{*}{\textbf{Method}} & \multicolumn{3}{c||}{\textbf{DLR}} & \multicolumn{3}{c|}{\textbf{MF}}\\ \cline{2-7}
                        & $gmean$ & $FDR$ & $FAR$ & $gmean$ & $FDR$ & $FAR$  \\ \hline
$HMM1$     &0	    &0	   &0.001 &0.092 &0.016 &0.005  \\ \hline
$HMM2$     &0	    &0	   &0.0003 &0	    &0	   &0.002 \\ \hline \hline
$XHMM1$           &0.854	&0.822 &0.096  &0.290 &0.094	&0.024 \\ \hline
$XHMM2$           &0.784	&0.965 &0.360  &\textbf{0.810} &0.978	&0.298 \\ \hline
$XHMM3$           &\textbf{0.925} &0.893 &0.030 &0.516	&0.285 &0.059 \\ \hline
$HMM_{NormOut}$  &0.326	&0.500 &0.731  &0.515 &0.399	&0.244 \\ \hline
$OCNN$           &0.380 &0.959 &0.846 &0.308 &0.736 &0.867 \\ \hline 
$OSVM$           &0.163 &0.117 &0.394 &0.652 &0.879 &0.508 \\ \hline

\end{tabular}
\caption{Performance of Fall Detection methods ($4$ states). For XHMM3 (\#states=\#labelled activities + $1$ state for unseen fall).}
\label{tab:exp1b}
\end{table}

\begin{figure}[!ht]
\vspace{-10mm}
\hspace*{-4cm}
  \begin{subfigure}[b]{0.5\textwidth}
  \centering
  \includegraphics[trim=0cm 7cm 0cm 7cm, clip,width=9.5cm,height=8cm]{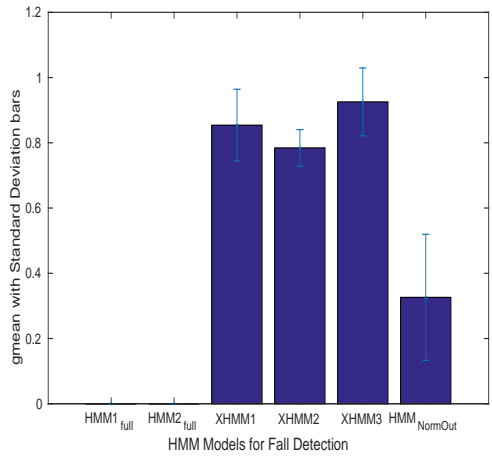} 
  \vspace{-15mm}
  \caption{DLR dataset}
  \label{fig:barDLR}
  \end{subfigure}
\hspace*{24mm}
  \begin{subfigure}[b]{0.5\textwidth}
  \centering
  \includegraphics[trim=0cm 7cm 0cm 7cm, clip,width=9.5cm,height=8cm]{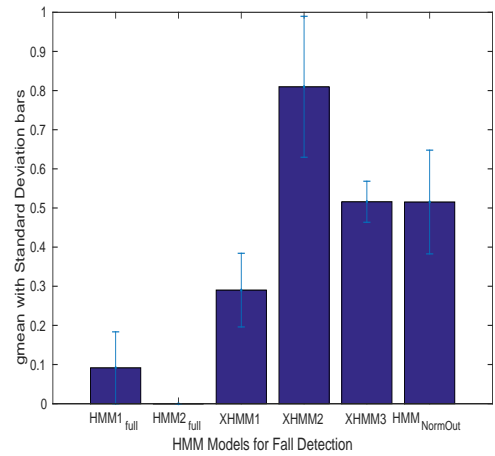} 
  \vspace{-15mm}
  \caption{MF dataset}
  \label{fig:barMF}
  \end{subfigure}
  \caption{$gmean$ with error bars across all subjects for DLR and MF datasets} 
  \label{fig:bar}
\end{figure}

To understand the statistical stability of the proposed methods, we plot the mean values of $gmean$ along with error bars (see Figure \ref{fig:bar}) representing standard deviation. Figure \ref{fig:bar} shows that for both the DLR and MF dataset, all the proposed XHMM methods outperform $HMM1$, $HMM2$ and $HMM_{NormOut}$.
Due to skewed distribution of falls in both the datasets, the standard deviation for the $gmean$ could be higher because a small number of misclassifications can vary the $gmean$ greatly. This experiment shows that training HMMs on full `normal' data for detecting unseen falls, and setting a threshold as the maximum of negative log-likelihood on training sequences is not the right approach and better models can be built when outliers from the `normal' datasets are removed and covariances of the X-Factor based HMMs are optimized.

\subsection{Feature Selection}

Selecting relevant features from a large set of features extracted from wearable sensors have shown to improve results for activity recognition \cite{Zhang:2011}. A major challenge in performing feature selection in the proposed problem of fall detection is that the fall data is not available during the training time; therefore, relevant features are to be selected from the non-fall data. We used the \textit{RELIEF-F} feature selection method \cite{Zhang:2011} for our task. \textit{RELIEF-F} computes a weight for each feature in terms of how well they distinguish between the data points of the same and different classes that are near to each other. This method provides a ranking of features in order of their merit for classification. We choose the top $10$ and top $20$ features and train the fall detection models discussed earlier with these reduced sets of features to study their effect on identifying unseen falls. The top selected features are mostly the mean, maximum, minimum, standard deviation, correlation, percentile and Signal Magnitude Area (see Table \ref{tab:selected_features}). Tables \ref{tab:dlrexp2} and \ref{tab:mfexp2} show that for both the DLR and MF datasets, reducing the number of features to $20$ from $31$ decrease the performance of $XHMM1$ and $XHMM3$ but increase the performance of $XHMM2$ and $HMM_{NormOut}$. When the number of features are reduced to the top $10$, the performance of all the classifiers deteriorates for the DLR and MF dataset (except for $XHMM3$). 
$OCNN$ and $OSVM$ performs worse in comparison to the XHMM methods. 
The degradation of performance can arise because feature selection is based on the normal activities only, instead of based on both falls and normal activites. This experiment shows that feature selection can improve the performance of the proposed XHMM methods.

 \begin{table}[!ht]
 \vspace{-10mm}
\begin{tabular}{|l|l|l|}\hline
\multirow{2}{*}{\textbf{Datasets}} & \multicolumn{2}{c|}{\cellcolor{Gray}\textbf{Top Ranked Features}}\\ \cline{2-3}
                                      & \textbf{Rank $1-10$} & \textbf{Rank $11-20$} \\ \hline
                            DLR       & $f_3$,$f_4$,$f_{23}$,$f_5$,$f_{19}$,       & $f_7$,$f_8$,$f_{15}$,$f_{22}$,$f_{30}$, \\ 
                                      & $f_{14}$,$f_9$,$f_{20}$,$f_{10}$,$f_{13}$  & $f_{18}$,$f_{31}$,$f_6$,$f_{29}$,$f_{11}$\\ \hline \hline
                            MF        & $f_{2}$,$f_{29}$,$f_{31}$,$f_{30}$,$f_3$,  & $f_9$,$f_4$,$f_8$,$f_{17}$,$f_{20}$, \\ 
                                      & $f_{11}$,$f_{19}$,$f_{13}$,$f_{22}$,$f_7$  & $f_{18}$,$f_5$,$f_6$,$f_{23}$,$f_{12}$\\ \hline 
   \end{tabular}
   \caption{Top $10/20$ ranked features. Compare with Table \ref{tab:features}}
   \label{tab:selected_features}
\end{table}

\begin{table}[!ht]
\begin{adjustwidth}{-0.8cm}{}
\centering
\begin{tabular}{| >{\centering\arraybackslash}m{2.1cm}| >{\centering\arraybackslash}m{0.9cm}| >{\centering\arraybackslash}m{0.65cm}| >{\centering\arraybackslash}m{0.7cm}|| >{\centering\arraybackslash}m{0.9cm}| >{\centering\arraybackslash}m{0.65cm}| >{\centering\arraybackslash}m{0.65cm}|} \hline
\multirow{2}{*}{\textbf{Method}} & \multicolumn{3}{c||}{\textbf{20 Features}} & \multicolumn{3}{c|}{\textbf{10 Features}}\\ \cline{2-7}
                        & $gmean$ & $FDR$ & $FAR$ & $gmean$ & $FDR$ & $FAR$  \\ \hline
  $HMM1$& 0     & 0     & 0     & 0.080 & 0.045 & 0 \\ \hline                        
  $HMM2$& 0     & 0     & 0.0001     & 0     & 0     & 0 \\ \hline \hline                       
  $XHMM1$      & 0.415 & 0.271 & 0.018 & 0.192 & 0.107 & 0.042\\ \hline                        
  $XHMM2$      & \textbf{0.852} & 0.933 & 0.213 & \textbf{0.832} & 0.933 & 0.248\\ \hline                        
  $XHMM3$      & 0.425 & 0.288 & 0.063 & 0.333 & 0.209 & 0.079\\ \hline                      
$HMM_{NormOut}$& 0.786 & 0.921 & 0.317 & 0.771 & 0.783 & 0.217\\ \hline
  $OCNN$	& 0.368 & 0.926 & 0.851 & 0.420 & 0.879 & 0.783 \\ \hline
  $OSVM$	& 0.237 & 0.203 & 0.501 & 0.053 & 0.039 & 0.553 \\ \hline
\end{tabular}
\caption{Performance of Fall Detection methods on reduced features for DLR dataset (Compare with Tables \ref{tab:exp1b})}
\label{tab:dlrexp2}
\end{adjustwidth}
\end{table}
 
\begin{table}[!ht]
\begin{adjustwidth}{-0.8cm}{}
\centering
\begin{tabular}{| >{\centering\arraybackslash}m{2.1cm}| >{\centering\arraybackslash}m{0.9cm}| >{\centering\arraybackslash}m{0.65cm}| >{\centering\arraybackslash}m{0.7cm}|| >{\centering\arraybackslash}m{0.9cm}| >{\centering\arraybackslash}m{0.65cm}| >{\centering\arraybackslash}m{0.65cm}|} \hline
\multirow{2}{*}{\textbf{Method}} & \multicolumn{3}{c||}{\textbf{20 Features}} & \multicolumn{3}{c|}{\textbf{10 Features}}\\ \cline{2-7}
                        & $gmean$ & $FDR$ & $FAR$ & $gmean$ & $FDR$ & $FAR$  \\ \hline
  $HMM1$& 0.093 & 0.020 & 0.007 & 0     & 0     & 0.005 \\ \hline                        
  $HMM2$& 0.106 & 0.022 & 0.002 & 0     & 0     & 0.005\\ \hline \hline                       
  $XHMM1$      & 0.051 & 0.008 & 0.004 & 0.046 & 0.006 & 0.005\\ \hline                        
  $XHMM2$      & \textbf{0.829} & 0.957 & 0.239 & \textbf{0.785} & 0.763 & 0.185\\ \hline                        
  $XHMM3$      & 0.531 & 0.333 & 0.110 & 0.685 & 0.542 & 0.109\\ \hline                      
$HMM_{NormOut}$& 0.759 & 0.774 & 0.163 & 0.566 & 0.453 & 0.127\\ \hline
  $OCNN$	& 0.303 & 0.686 & 0.861 & 0.324 & 0.695 & 0.842\\ \hline 
  $OSVM$	& 0.579 & 0.717 & 0.516 & 0.658 & 0.933 & 0.508\\ \hline
\end{tabular}
\caption{Performance of Fall Detection methods on reduced features for MF dataset (Compare with Tables \ref{tab:exp1b})}
\label{tab:mfexp2}
\end{adjustwidth}
\end{table}

\subsection{Training with fall data}
In this experiment, we compare several supervised classification algorithms for fall detection under two scenarios (a) when full data for falls is available, and (b) when small amount of fall data is available during training and is gradually increased. The latter experiment simulates a scenario when we may have few fall data to begin with. We simulate this scenario by supplying a controlled amount of fall data during the training phase and train the supervised classifiers by randomly choosing $1, 2, 4, 6, 8, 10, 25,$ and $50$ falls samples from the full fall data. To avoid classification bias due to random choice of fall data, we run this experiment $10$ times (per LOOCV fold) and report the average value of the performance metrics. 
We use supervised version of the $XHMM$s presented earlier. $HMM1_{sup}$ is similar to $XHMM1$, where each normal activity is modelled by a separate HMM by utilizing full `normal' data for each activity; however, due to the presence of fall data a separate HMM is trained for fall events. $HMM2_{sup}$ is similar to $XHMM2$, where the full `normal' activities are modelled by a general HMM and a separate HMM is trained to model falls. $HMM3_{sup}$ is similar to $XHMM3$; however, in this case a state representing `actual' fall activity is added in the HMM and its parameters are computed from the labelled fall data. The other two supervised classifiers we use are Random Forest ($RF$) and Support Vector Machine ($SVM$). The ensemble size in $RF$ is set to $200$, where each decision (or split) in each tree is based on a single, randomly selected feature \cite{Stone2015Fall}. For $SVM$ classifier, a Gaussian kernel is used with width equals to $10$. 

\begin{table}[!ht]
\vspace{2mm}
\hspace{-6mm}
\centering
\begin{tabular}{| >{\centering\arraybackslash}m{1.6cm}| >{\centering\arraybackslash}m{0.9cm}| >{\centering\arraybackslash}m{0.65cm}| >{\centering\arraybackslash}m{0.65cm}|| >{\centering\arraybackslash}m{0.9cm}| >{\centering\arraybackslash}m{0.65cm}| >{\centering\arraybackslash}m{0.65cm}|} \hline
\multirow{2}{*}{\textbf{Method}} & \multicolumn{3}{c||}{\textbf{DLR}} & \multicolumn{3}{c|}{\textbf{MF}}\\ \cline{2-7}
                        & $gmean$ & $FDR$ & $FAR$ & $gmean$ & $FDR$ & $FAR$  \\ \hline
  $HMM1_{sup}$ &0.768	&0.719	&0.054 &0.489	&0.259	&0.038\\ \hline                        
  $HMM2_{sup}$ &0.601	&0.533	&0.087 &0.925	&0.939	&0.084\\ \hline                        
  $HMM3_{sup}$ &\textbf{0.938}	&0.908	&0.021 &\textbf{0.969}	&0.988	&0.045\\ \hline     
  $RF$         & 0.622 & 0.496 & 0.001 & 0.962 & 0.937 & 0.012 \\ \hline
  $SVM$        &\textbf{0.929} & 0.885 & 0.015 & \textbf{0.985} & 0.994 & 0.025 \\ \hline
\end{tabular}
\caption{Supervised Fall Detection with full training data for falls and all normal activities (Compare with Table \ref{tab:exp1b}).}
\label{tab:exp3a}
\end{table}

\begin{figure}
\begin{adjustwidth}{-3cm}{}\centering
\vspace{-15mm}
  \begin{subfigure}[b]{0.6\textwidth}
  \centering
  \includegraphics[trim=4cm 8cm 2cm 9cm, clip,width=1.8 \textwidth]{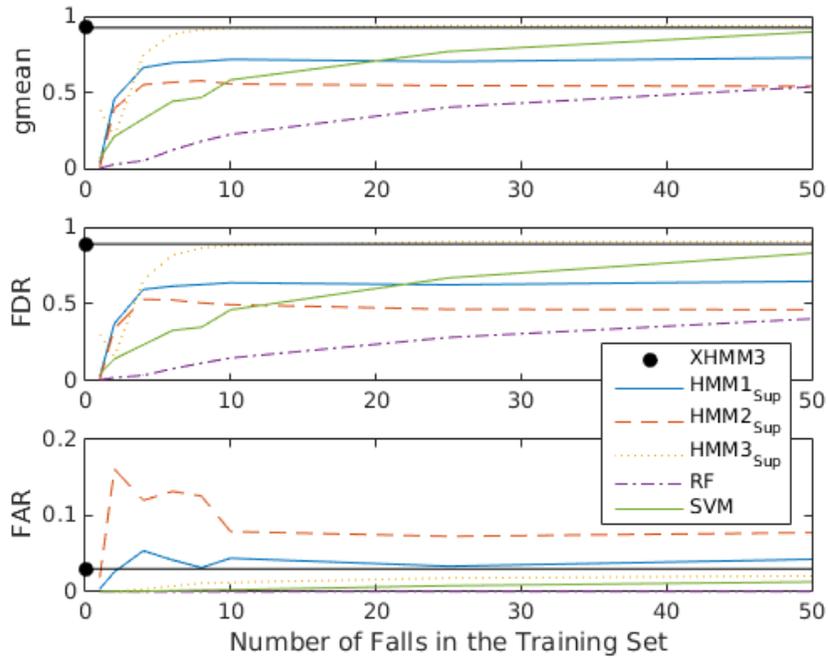} 
  \vspace{-5mm}
  \caption{DLR dataset}
  \label{fig:dlrexp4}
  \end{subfigure}\\
  \begin{subfigure}[b]{0.6\textwidth}
  \centering
  \includegraphics[trim=4cm 8cm 2cm 9cm, clip,width=1.8 \textwidth]{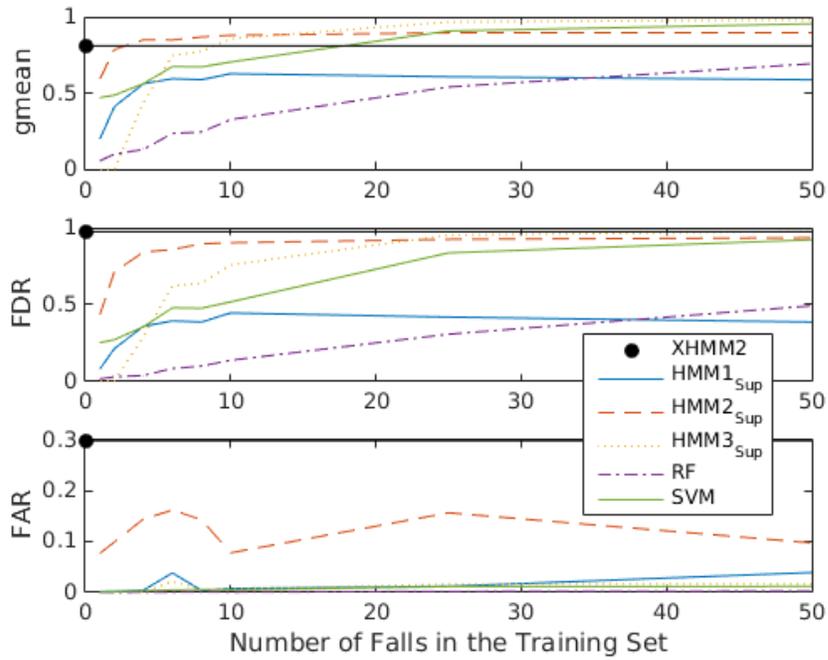} 
  \vspace{-5mm}
  \caption{MF dataset}
  \label{fig:mfexp4}
  \end{subfigure}
  \caption{Effect of varying the amount of fall data in supervised learning. Two best performing X-Factor approaches are shown on the y-axis corresponding to zero training data (compare with Table \ref{tab:exp1b}).} 
\end{adjustwidth}
\end{figure}

Table \ref{tab:exp3a} shows the LOOCV results for both the datasets when full training data is available for falls and all the normal activities. For the MF dataset, the performance improvements in all the $XHMM$ based classifiers in comparison to their counterparts that are trained in the absence of falls. For the DLR dataset, performance of $HMM1_{sup}$ and $HMM2_{sup}$ is worse than when no training data for falls is used, whereas $HMM3_{sup}$ show improvement with equivalent performance as $SVM$. The $RF$ classifier gives intermediate results.
Figures \ref{fig:dlrexp4} and \ref{fig:mfexp4} show the performance of supervised classifiers when the number of fall data is gradually increased during the training phase for the DLR and MF dataset. All the supervised classifiers perform worse when the training data for falls is very small. Figure \ref{fig:dlrexp4} shows that as the number of samples in the training data for falls increase, $HMM3_{sup}$ and $SVM$ starts to perform better than other classifiers but provides equivalent performance to $XHMM3$ (shown by $\bullet$ on the y-axis representing no training data for falls). The performance of $XHMM3$, which requires no fall data for training is much better than its supervised counterpart ($HMM3_{sup}$) when a small number of  training samples for falls is available. Figure \ref{fig:mfexp4} shows that the performance of $HMM2_{sup}$ starts to improve when some fall data are added in the training set for MF dataset, whereas other classifiers perform worse with limited training samples for falls. $XHMM2$ and $HMM2_{sup}$ with small number of training samples for falls show comparable performance. As the number of fall samples increase in the training set, $HMM3_{sup}$ and $SVM$ outperform other methods. 

Both the experiments on the DLR and MF datasets suggest that the performance of supervised classifiers improve as the number of fall samples increase in the training set. However, when they are trained on very limited fall data, their performance is worse in comparison to the proposed models that did not observe falls before.
 The results from the study of Stone and Skubic \cite{Stone2015Fall} show that only $9$ actual falls were obtained over a combined nine years of continuous activity data in a realistic setting,  which highlights the rarity of fall occurrence and consequently the difficulty in training supervised classifiers on abundant fall data. 
 Moreover, supervised methods \textit{cannot} handle training the classifiers in the absence of falls, whereas the proposed X-factor approaches can learn in the absence of training data for falls and identify them with high $gmean$ and $FDR$.

\subsection{Are outliers representative of proxy for falls?}
\label{sec:exp4}
Section \ref{sec:threshold} assumes that the outlier sequences present in the normal activities can be used as a proxy for falls to estimate the parameters $\xi$. We conduct an experiment to  evaluate the validity of this assumption. We use the supervised HMMs ($HMM1_{sup}$ and $HMM2_{sup}$), with the only difference that they are trained on `non-fall' activity (i.e. obtained after removing outliers from the normal data) and falls. During the testing phase we present the `outliers' to the classifier instead of normal and fall data. The idea is that some of the outliers  that are rejected by the normal activities will be classified as falls as they differ from the normal activities or the general non-fall concept due to inadvertent sensor artifacts. 

\paragraph*{$HMM1_{sup}$}
When using $HMM1_{sup}$, for the DLR dataset, the outliers of normal activities `Jumping' and `Running' are most of the time classified as `Falls', the outliers from the activities `Walking' and `Lying' are sometimes classified as falls, whereas outliers from `Sitting' and `Standing' are mostly classified as non-falls. This provides evidence that some of the short term dynamic activities can have variations and may not be identified correctly in their respective classes. Similar experiments on the MF dataset show that only the step-in car activity's outliers are classified as falls and the rest of the outliers of other `non-fall' activities are classified as non-falls.

\paragraph*{$HMM2_{sup}$}
When using $HMM2_{sup}$, for the MF dataset, the outliers are mostly classified as falls and for the DLR dataset, they are classified as non-falls. 

Based on the above experiments, we can conclude that in the absence of fall data during training, rejected outliers from the normal activities can be used as a proxy for falls, provided they are very different from the samples of normal activities or the general concept of normal activity. 
However, it is to be noted that since these rejected outliers are not actual falls and only some of them are similar to falls.

\section{Conclusions and Future Work}
The lack of sufficient data for falls can adversely affect the performance of supervised fall detection classifiers. Moreover, the supervised classification methods cannot handle the realistic scenario when no training data for falls is available. In this paper, we present three `X-factor' HMM based fall detection approaches that learn only from the normal activities captured from a body-worn sensor. To tackle the issue of no training data for falls, we introduced a new cross-validation method based on the inter-quartile range of log-likelihoods on the training data that rejects spurious data from the normal activities, treats them as proxies for unseen falls and helps in optimizing the model parameter. The results showed that two of the XHMM methods show high detection rates for falls in person and placement of sensor independent manner. We showed that the traditional method of thresholding with HMM on full normal data set as maximum of negative log-likelihood to identify unseen falls is not the right approach for this problem.  We also showed that supervised classifiers performed poorly with few training samples for falls, whereas in comparison the proposed methods show high performance in the absence of training data for falls. An important extension of the proposed techniques is the realization of an online fall detection system, which can begin with X-factor models as initial representative model for unseen falls and incrementally adapts its parameters as it starts identifying some falls. 
\section*{References}
\bibliography{references}

\begin{thebibliography}{10}
\expandafter\ifx\csname url\endcsname\relax
  \def\url#1{\texttt{#1}}\fi
\expandafter\ifx\csname urlprefix\endcsname\relax\def\urlprefix{URL }\fi
\expandafter\ifx\csname href\endcsname\relax
  \def\href#1#2{#2} \def\path#1{#1}\fi

\bibitem{Acampora2013}
G.~Acampora, D.~Cook, P.~Rashidi, A.~Vasilakos, A survey on ambient
  intelligence in healthcare, Proceedings of the IEEE 101~(12) (2013)
  2470--2494.

\bibitem{igual2013challenges}
R.~Igual, C.~Medrano, I.~Plaza, Challenges, issues and trends in fall detection
  systems, BioMedical Engineering OnLine 12~(1) (2013) 1--24.

\bibitem{website:cdc}
CDC, Falls in nursing homes,
  \url{http://www.cdc.gov/HomeandRecreationalSafety/Falls/nursing.html},
  accessed on $15^{th}$ June 2016.

\bibitem{Debard2012Camera}
G.~Debard, P.~Karsmakers, M.~Deschodt, E.~Vlaeyen, E.~Dejaeger, K.~Milisen,
  T.~Goedemé, B.~Vanrumste, T.~Tuytelaars, Camera-based fall detection on real
  world data, in: Outdoor and Large-Scale Real-World Scene Analysis, Vol. 7474
  of Lecture Notes in Computer Science, Springer Berlin Heidelberg, 2012, pp.
  356--375.

\bibitem{Stone2015Fall}
E.~Stone, M.~Skubic, Fall detection in homes of older adults using the
  microsoft kinect, Biomedical and Health Informatics, IEEE Journal of 19~(1)
  (2015) 290--301.

\bibitem{kangas2012comparison}
M.~Kangas, I.~Vikman, L.~Nyberg, R.~Korpelainen, J.~Lindblom, T.~J{\"a}ms{\"a},
  Comparison of real-life accidental falls in older people with experimental
  falls in middle-aged test subjects, Gait \& posture 35~(3) (2012) 500--505.

\bibitem{kulic08}
D.~Kuli\'c, W.~Takano, Y.~Nakamura, Incremental learning, clustering and
  hierarchy formation of whole body motion patterns using adaptive hidden
  markov chains., The International Journal of Robotics Research 27~(7) (2008)
  761--784.

\bibitem{khan2014iwaal}
S.~S. Khan, M.~E. Karg, D.~Kuli\'c, J.~Hoey, X-factor {HMM}s for detecting
  falls in the absence of fall-specific training data, in: L.~P. et~al. (Ed.),
  Proceedings of the 6th International Work-conference on Ambient Assisted
  Living (IWAAL 2014), Vol. 8868, Springer International Publishing
  Switzerland, Belfast, U.K., 2014, pp. 1--9.

\bibitem{Mubashir:2013:SFD}
M.~Mubashir, L.~Shao, L.~Seed, A survey on fall detection: Principles and
  approaches, Neurocomput. 100 (2013) 144--152.

\bibitem{kwolek2014human}
B.~Kwolek, M.~Kepski, Human fall detection on embedded platform using depth
  maps and wireless accelerometer, Computer methods and programs in biomedicine
  117~(3) (2014) 489--501.

\bibitem{Bourke200884}
A.~Bourke, G.~Lyons, A threshold-based fall-detection algorithm using a
  bi-axial gyroscope sensor, Medical Engineering and Physics 30~(1) (2008) 84
  -- 90.

\bibitem{Khan:KER:2014}
S.~S. Khan, M.~G. Madden, One-class classification: taxonomy of study and
  review of techniques, The Knowledge Engineering Review 29 (2014) 345--374.

\bibitem{DBLP:conf/icarcv/ThomeM06}
N.~Thome, S.~Miguet, A hhmm-based approach for robust fall detection, in:
  ICARCV, IEEE, 2006, pp. 1--8.

\bibitem{Luo:Liu:Liu:2012}
X.~Luo, T.~Liu, J.~Liu, X.~Guo, G.~Wang, Design and implementation of a
  distributed fall detection system based on wireless sensor networks, EURASIP
  Journal on Wireless Communications and Networking 2012 (2012) 1--13.

\bibitem{Tokumitsu:2011}
M.~Tokumitsu, M.~Murakami, Y.~Ishida, An adaptive sensor network for home
  intrusion detection by human activity profiling, Artificial Life and Robotics
  16~(1) (2011) 36--39.

\bibitem{iet/conferences/chen:luo:2011}
H.~Cheng, L.~Haiyong, F.~Zhao, A fall detection algorithm based on pattern
  recognition and human posture analysis, in: IET International Conference on
  Communication Technology and Application, 2011, pp. 853--857.

\bibitem{hunag:li:irwin:2006}
T.~Zhang, J.~Wang, L.~Xu, P.~Liu, Fall detection by wearable sensor and
  one-class svm algorithm, in: Intelligent Computing in Signal Processing and
  Pattern Recognition, Vol. 345, Springer Berlin Heidelberg, 2006, pp.
  858--863.

\bibitem{Miao:Naqvi:2011}
M.~Yu, S.~Naqvi, A.~Rhuma, J.~Chambers, Fall detection in a smart room by using
  a fuzzy one class support vector machine and imperfect training data, in:
  ICASSP, 2011, pp. 1833--1836.

\bibitem{Popescu2009}
M.~Popescu, A.~Mahnot, Acoustic fall detection using one-class classifiers, in:
  Annual International Conference of the IEEE EMBC, 2009, pp. 3505--3508.

\bibitem{medrano2014detecting}
C.~Medrano, R.~Igual, I.~Plaza, M.~Castro, Detecting falls as novelties in
  acceleration patterns acquired with smartphones, PloS one 9~(4) (2014)
  e94811.

\bibitem{rabiner}
L.~Rabiner, A tutorial on hidden markov models and selected applications in
  speech recognition, Proceedings of the IEEE 77~(2) (1989) 257--286.

\bibitem{Manninis100201154}
A.~Mannini, A.~M. Sabatini, Machine learning methods for classifying human
  physical activity from on-body accelerometers, Sensors 10~(2) (2010)
  1154--1175.

\bibitem{quinn}
J.~A. Quinn, C.~K. Williams, N.~McIntosh, Factorial switching linear dynamical
  systems applied to physiological condition monitoring, IEEE Transactions on
  PAMI 31~(9) (2009) 1537--1551.

\bibitem{Khan:2012:TDU:2370216.2370444}
S.~S. Khan, M.~E. Karg, J.~Hoey, D.~Kuli\'c, Towards the detection of unusual
  temporal events during activities using hmms, in: SAGAWARE - Proceedings of
  the 2012 ACM Conference on Ubiquitous Computing, UbiComp '12, ACM, 2012, pp.
  1075--1084.

\bibitem{smyth1994markov}
P.~Smyth, Markov monitoring with unknown states, Selected Areas in
  Communications, IEEE Journal on 12~(9) (1994) 1600--1612.

\bibitem{dlr65511}
M.~J.~V. Nadales, Recognition of human motion related activities from sensors,
  Master's thesis, University of Malaga and German Aerospace Cener (2010).

\bibitem{mobifall}
G.~Vavoulas, M.~Pediaditis, E.~Spanakis, M.~Tsiknakis, The mobifall dataset: An
  initial evaluation of fall detection algorithms using smartphones, in:
  Bioinformatics and Bioengineering (BIBE), 2013 IEEE 13th International
  Conference on, 2013, pp. 1--4.

\bibitem{delavega2013}
L.~G.~M. de~la Vega, S.~Raghuraman, A.~Balasubramanian, B.~Prabhakaran,
  Exploring unconstrained mobile sensor based human activity recognition, in:
  3rd International Workshop on Mobile Sensing, 2013.

\bibitem{winter2009biomechanics}
D.~A. Winter, Biomechanics and motor control of human movement, John Wiley \&
  Sons, 2009.

\bibitem{Ravi:2005}
N.~Ravi, N.~Dandekar, P.~Mysore, M.~L. Littman, Activity recognition from
  accelerometer data, in: Proceedings of the 17th conference on Innovative
  applications of artificial intelligence - Volume 3, IAAI'05, AAAI Press,
  2005, pp. 1541--1546.

\bibitem{huynh2005analyzing}
T.~Huynh, B.~Schiele, Analyzing features for activity recognition, in:
  Proceedings of the 2005 joint conference on Smart objects and ambient
  intelligence: innovative context-aware services: usages and technologies,
  ACM, 2005, pp. 159--163.

\bibitem{cruz2013features}
N.~Cruz-Silva, J.~Mendes-Moreira, P.~Menezes, Features selection for human
  activity recognition with iphone inertial sensors, in: Portuguese Conference
  on Artificial Intelligence, 2013.

\bibitem{Ermes2008}
M.~Ermes, J.~Parkka, L.~Cluitmans, Advancing from offline to online activity
  recognition with wearable sensors, in: 2008. 30th Annual International
  Conference EMBS, 2008, pp. 4451--4454.

\bibitem{khan2008accelerometer}
A.~M. Khan, Y.-K. Lee, T.-S. Kim, Accelerometer signal-based human activity
  recognition using augmented autoregressive model coefficients and artificial
  neural nets, in: Engineering in Medicine and Biology Society, 2008. EMBS
  2008. 30th Annual International Conference of the IEEE, IEEE, 2008, pp.
  5172--5175.

\bibitem{DBLP:conf/pervasive/BaoI04}
L.~Bao, S.~S. Intille, Activity recognition from user-annotated acceleration
  data, in: A.~Ferscha, F.~Mattern (Eds.), Pervasive, Vol. 3001 of Lecture
  Notes in Computer Science, Springer, 2004, pp. 1--17.

\bibitem{cleland2013optimal}
I.~Cleland, B.~Kikhia, C.~Nugent, A.~Boytsov, J.~Hallberg, K.~Synnes,
  S.~McClean, D.~Finlay, Optimal placement of accelerometers for the detection
  of everyday activities, Sensors 13~(7) (2013) 9183--9200.

\bibitem{karg2014clinical}
M.~Karg, W.~Seiberl, F.~Kreuzpointner, J.~P. Haas, D.~Kulić, Clinical gait
  analysis: Comparing explicit state duration hmms using a reference-based
  index, IEEE Transactions on Neural Systems and Rehabilitation Engineering
  23~(2) (2015) 319--331.

\bibitem{he2009activity}
Z.~He, L.~Jin, Activity recognition from acceleration data based on discrete
  consine transform and svm, in: SMC, IEEE, 2009, pp. 5041--5044.

\bibitem{kubat1997addressing}
M.~Kubat, S.~Matwin, Addressing the curse of imbalanced training sets:
  one-sided selection, in: ICML, Vol.~97, 1997, pp. 179--186.

\bibitem{khan2010kernels}
S.~S. Khan, Kernels for one-class nearest neighbour classification and
  comparison of chemical spectral data, Master's thesis, College of Engineering
  and Informatics, National University of Ireland (2010).

\bibitem{website:fitcsvm}
MATLAB, fitcsvm, \url{https://www.mathworks.com/help/stats/fitcsvm.html},
  accessed on $19^{th}$ September 2016.

\bibitem{Zhang:2011}
M.~Zhang, A.~A. Sawchuk, A feature selection-based framework for human activity
  recognition using wearable multimodal sensors, in: Proceedings of the 6th
  International Conference on Body Area Networks, BodyNets '11, ICST, Brussels,
  Belgium, Belgium, 2011, pp. 92--98.

\end{thebibliography}
\end{document}